\documentclass[
twocolumn,
]{ceurart}

\usepackage{listings}
\usepackage[a4paper]{geometry} 
\usepackage{graphicx}
\usepackage{xurl} 
\usepackage[italian,english]{babel}
\usepackage{latexsym} 
\usepackage{xcolor}
\usepackage{verbatim}
\usepackage{amsmath}
\usepackage{caption}

\newcommand{\sg}[1]{\textcolor{blue}{#1-sg}}
\newcommand{\pl}[1]{\textcolor{red}{#1-pl}}

\newcommand{\ul}[1]{\underline{#1}}
\newcommand{\od}{\emph{Od}}
\newcommand{\caus}{\emph{Caus}}

\usepackage{graphicx}
\usepackage{caption}
\usepackage{subcaption}

\begin{document}

\copyrightyear{2022}
\copyrightclause{Copyright for this paper by its authors.
  Use permitted under Creative Commons License Attribution 4.0
  International (CC BY 4.0).}

\conference{CLiC-it 2024: 10th Italian Conference on Computational Linguistics, Dec 04 — 06, 2024, Pisa, Italy}

\title{Exploring Italian sentence embeddings properties \\ through multi-tasking 
}

\author[1]{Vivi Nastase}[email=vivi.a.nastase@gmail.com]
\cormark[1]
\author[1]{Giuseppe Samo}[email=giuseppe.samo@idiap.ch]
\author[1,2]{Chunyang Jiang}[email=chunyang.jiang42@gmail.com]
\author[1,2]{Paola Merlo}[email=Paola.Merlo@unige.ch]
\address[1]{Idiap Research Institute, Martigny, Switzerland}
\address[2]{University of Geneva, Geneva, Switzerland}

\cortext[1]{Corresponding author.}

\begin{abstract}
We investigate to what degree existing LLMs encode abstract linguistic information in Italian in a multi-task setting. We exploit curated synthetic data on a large scale -- several Blackbird Language Matrices (BLMs) problems in Italian -- and use them to study how sentence representations built using pre-trained language models encode specific syntactic and semantic information. We use a two-level architecture to model separately a compression of the sentence embeddings into a representation that contains relevant information for a task, and a BLM task. We then investigate whether we can obtain compressed sentence representations that encode syntactic and semantic information relevant to several BLM tasks. While we expected that the sentence structure -- in terms of sequence of phrases/chunks -- and chunk properties could be shared across tasks, performance and error analysis show that the clues for the different tasks are encoded in different manners in the sentence embeddings, suggesting that abstract linguistic notions such as constituents or thematic roles does not seem to be present in the pretrained sentence embeddings.\\

\noindent L'obiettivo di questo lavoro è indagare fino a che punto gli attuali LLM  apprendono rappresentazioni linguistiche astratte in configurazioni multitask. Utilizzando dati sintetici curati su larga scala di vari problemi BLM in italiano, studiamo come le rappresentazioni di frasi costruite da modelli di linguaggio pre-addestrati codifichino le informazioni semantiche e sintattiche. Abbiamo utilizzato un'architettura a due livelli per modellare separatamente, da un lato, la compressione degli embeddings delle frasi di input in una rappresentazione che contiene informazioni rilevanti per i tasks BLM e, dall'altro lato, il BLM stesso. Abbiamo poi verificato se fosse possibile ottenere rappresentazioni compresse di frasi che codificano informazioni sintattiche e semantiche rilevanti per i diversi tasks BLM. Contrariamente alla predizione che la struttura della frase - in termini di sequenza di frasi/chunks - e le proprietà dei chunk possano essere condivise tra i vari tasks, i risultati e l'analisi degli errori mostrano che gli indizi per i diversi task sono codificati in modo diverso negli embeddings delle frasi. Questo risultato suggerisce che nozioni linguistiche astratte come i costituenti o i ruoli tematici non vi sembrano essere presenti.

\end{abstract}


\maketitle

\section{Introduction}

Driven by increasing computational scale and progress in deep learning techniques, NLP models can rival human capabilities on established benchmarks. New benchmarks, then, that capture deeper levels of language understanding must be created and analysed \cite{ruder2021benchmarking}.

Blackbird's Language Matrices (BLM) \cite{merlo2023} is a recent task inspired by visual tests of analytic intelligence (Raven Progressive Matrices/RPMs, \cite{raven1938}). 
The BLM tasks have cast light on  whether the correct predictions in previously studied linguistic problems, e.g. number agreement or verb alternations, stem from sentence embeddings that encode deeper linguistic information, such as syntactic structure and semantic properties of phrases \cite{an-etal-2023-blm,nastase-merlo-2023-grammatical,nastase2024identifiable}. We  found that higher-level information -- syntactic structure and argument structure -- can be assembled from the information encoded in the sentence embeddings. This, however, may not be due to a deeper understanding of such information encoded by LLMs, but rather because of useful surface indicators \cite{lenci2023NLU}.


In this paper, we adopt BLMs to investigate whether current pretrained models encode abstract linguistic notions, such as constituents, and are able to do so in a manner that comprises both functional elements, such as pronouns, demonstratives and lexical elements, such as nominal constituents.

We concentrate on Italian, and study several grammatical problems whose solutions can theoretically help each other, in a multi-task setting.
We adopt a two-level architecture developed specifically to model  what we know about how humans solve puzzles similar to BLMs \cite{carpenter-ea1990}. Level 1 aims to obtain compressed sentence representations that capture information about constituents and their properties; level 2 uses the compressed sentence representations to solve a BLM problem. This architecture provides a tool to study how LLMs encode different types of syntactic and semantic information.

We make two contributions: (i) an initial core BLM dataset for Italian that covers linguistic problems of different nature; (ii)
single and multi-task experiments that provide new insights into the information encoded by LLMs. The datasets are available at \url{https://www.idiap.ch/dataset/(blm-agri|blm-causi|blm-odi)} and the code at \url{https://github.com/CLCL-Geneva/BLM-SNFDisentangling}.


\section{Related Work}

Multi-task learning  has been popular in improving NLP systems' performance by using knowledge shared across multiple tasks \cite{zhang-etal-2023-survey}. 

Multi-task learning architectures include parallel, hierarchical, and modular designs \cite{chen2021multi}. Parallel architectures share intermediate layers across tasks, conducive to efficient knowledge transfer \cite{ruder2017overview}. Hierarchical architectures capture task dependencies by layering task-specific modules on shared bases. Modular approaches selectively share components among tasks to balance between generalisation and task-specific optimisation \cite{pfeiffer2023modular}. These training strategies are not mutually exclusive and can be combined.

Multi-task learning can be used efficiently in resource-constrained environments, to counter data scarcity and overfitting: aggregating training data and sharing parameters across related tasks acts as a form of data augmentation \cite{standley2020tasks}. 

Effective multi-task learning depends on the relatedness of the tasks involved. Tasks that are similar or have similar objectives tend to benefit more from shared representations. This observation has been used in various NLP tasks, including named entity recognition \cite{zhou-etal-2021-end}, text generation\cite{hu-etal-2022-mocha}, and machine translation \cite {wang-etal-2020-multi}, among others.  Selecting related tasks that contribute positively to the shared model's training is important and remains an active area of research \cite{zhang-etal-2023-survey}. 

Pretrained large language models exhibit general-purpose abilities and knowledge, with high results with little or no fine-tuning on downstream tasks \cite{devlin-etal-2019-bert,clark2020electra}. We can then regard these language models as the results of "multi-task" learning, and our aim here is to test whether sentence embeddings obtained from these models encode syntactic and semantic information consistently, such that different BLM problems that rely on similar linguistic information draw on the same clues from these representations. In particular, we will use BLM tasks on subject-verb agreement -- which relies on chunk structure and the chunks' grammatical number properties -- and on verb alternations -- which relies on chunk structure and the chunks' semantic role properties -- to test whether chunk structure is encoded in a manner that allows for it to be shared by the two tasks.

\section{The BLM task and the BLM Italian datasets}
\label{sec:data}


Raven's progressive matrices are multiple-choice completion IQ tests, whose solution requires discovering underlying generative rules of a sequence of images \cite{raven1938}.  

\begin{figure}
\begin{minipage}{0.48\textwidth}
\small
\begin{tabular}{llll} 
\multicolumn{4}{l}{{\bf BLM agreement problem} (BLM-AgrI)} \\
\hline
\multicolumn{4}{c}{\sc Context Template}\\
\hline
\sg{NP}& \sg{PP1}& & \sg{VP}  \\
\pl{NP} & \sg{PP1}& & \pl{VP}  \\
\sg{NP}& \pl{PP1} & & \sg{VP}  \\
\pl{NP} & \pl{PP1} & & \pl{VP}  \\
\sg{NP}& \sg{PP1}& \sg{PP2} &\sg{VP}  \\
\pl{NP} & \sg{PP1}  & \sg{PP2} & \pl{VP}  \\
\sg{NP}   & \pl{PP1} & \sg{PP2} &  \sg{VP}  \\
\end{tabular}

  \begin{tabular}{llllr} \hline
 \multicolumn{5}{c}{{\sc Answer set} } \\ \hline
\rowcolor{lightgray!40} 
\pl{NP} & \pl{PP1} & \sg{PP2} & \pl{VP} & \textsc{Correct} \\
\pl{NP} & \pl{PP1} & et \sg{PP2} & \pl{VP} & Coord  \\ 
\pl{NP} & \pl{PP1} &          & \pl{VP} & WNA\\
\pl{NP} & \sg{PP1} & \sg{PP1} & \pl{VP} & WN1 \\
\pl{NP} & \pl{PP1} & \pl{PP2} & \pl{VP} & WN2 \\
\pl{NP} & \pl{PP1} & \pl{PP2} & \sg{VP} & AEV \\
\pl{NP} & \sg{PP1} & \pl{PP2} & \sg{VP} & AEN1 \\
\pl{NP} & \pl{PP1} & \sg{PP2} & \sg{VP} & AEN2 \\ \hline
 \end{tabular}
\end{minipage}
\caption{BLM instances for verb-subject agreement, with two attractors. We build candidate answers displaying one of two types of errors:
(i) \ul{sequence errors}: WNA= wrong nr. of attractors; WN1= wrong gram. nr. for 1$^{st}$ attractor noun (N1); WN2= wrong gram. nr. for 2$^{nd}$ attractor noun (N2); (ii) \ul{grammatical errors}: AEV=agreement error on the verb; AEN1=agreement error on N1; AEN2=agreement error on N2.}
\vspace{-5mm}
\label{fig:agr_templ}
\end{figure}

A similar task has been developed for linguistic problems, called Blackbird Language Matrices (BLMs) \cite{merlo2023}, as given in Figure \ref{fig:agr_templ}, which illustrates the template of a BLM agreement matrix. A BLM comprises a context and an answer set. The context is a sequence of sentences generated following the relevant rules of a given linguistic phenomenon under investigation and that this way implicitly illustrates these grammatical properties. This sequence also follows some extra-linguistic progression rules. 
Each context is paired with a set of candidate answers. The answer sets contain minimally contrastive examples built by corrupting some of the generating rules. 

The BLM Italian datasets consists of BLMs focused on the property of subject-verb agreement and two transitive-intransitive alternations: the change-of-state alternation and the object-drop alternation.

\subsection{BLM-AgrI -- subject-verb agreement in Italian}

The BLM-AgrI dataset is created by manually translating the seed French sentences \cite{an-etal-2023-blm} into Italian by a native  speaker, one of the authors, and then generating the full dataset following the same process of lexical augmentation and sentence shuffling among instances described in \cite{an-etal-2023-blm}. The internal nominal structure in these languages is very similar, so translations are almost parallel. An illustrative, simplified example for Italian is provided in Figure \ref{fig:matrix-example}, in the appendix.
The dataset comprises three subsets of increasing lexical complexity (called Type I, Type II and Type III) to test the ability of the system to handle item novelty.

\subsection{BLM-CausI and BLM-OdI}

While BLM-AgrI tests information about a formal grammatical property, agreement, the Causative (\caus) and Object-drop (\od\/) alternation datasets test lexical semantic properties of verbs, their ability to enter or not a causative alternation.
\caus~ represents the causative/inchoative alternation, where the object of the transitive verb bears the same semantic role (Patient) as the subject of the intransitive verb (\textit{L'artista ha aperto la finestra/La finestra si è aperta} `The artist opened the window'/`The window opened'). The transitive form of the verb has a causative meaning. In contrast, the subject in \od~ bears the same semantic role (Agent) in both the transitive and intransitive forms (\textit{L'artista dipingeva la finestra/L'artista dipingeva} `the artist painted the window'/`the artist painted') and the verb does not have a causative meaning \citep{Levin93,merlo2001automatic}.

\paragraph{BLM-CausI context and answers}

The context set of the verb alternation varies depending on the presence of one or two arguments and their attributes (agents, \textcolor{violet}{Ag}; patients, \textcolor{orange}{Pat}) and  the active (\textcolor{brown}{Akt}) and passive (\textcolor{teal}{Pass}) or passive voice of the verb. The non-linguistic factor that structures the sequence is an alternation every two items between a prepositional phrase introduced  by any preposition (e.g., \textit{in pochi secondi}, \textcolor{blue}{P}-NP) and a PP introduced by the agentive \textcolor{red}{da}-NP (e.g., \textit{dall'artista}, \textcolor{red}{da}-\textcolor{violet}{Ag}/\textcolor{red}{da}-\textcolor{orange}{Pat}).

The answer set is composed of one correct answer and contrastive wrong answers, all formed by  the same four elements: a verb, two nominal constituents and a prepositional phrase.  
Figure \ref{tab:templateBLM} shows the template.\footnote{Following BLM formal specifications \cite{merlo2023}, we build the errors representing violations of internal (\textit{I}), external (\textit{E}) and relational (\textit{R}) rules of the BLM, and their combination (e.g. \textit{IE} \textit{IER}, etc.). This information is used in the first part of the error acronym. The second part of the errors' label indicates the structure the sentence represent: intransitive (\textsc{Int}), passive (\textsc{Pass}), Transitive (\textsc{Trans}) or, in some cases, the NP introduced by the \textcolor{red}{\textit{da}} preposition (\textsc{WrBy)}.}

\paragraph{BLM-OdI Context and Answers}

The BLM for \od~ is the same as for \caus, but here 
the passive voice serves as a confounding element and one of the contrastive answers for \caus~ is, in fact, the correct answer here.

The template is also in Figure \ref{tab:templateBLM}. 
Due to the asymmetry between the two classes of verbs, the contexts of the BLMs minimally differ in the intransitive followed by \textcolor{blue}{P}-NP (sentence 7). The correct answer also varies across the two groups, although in both cases it is an intransitive form with a \textcolor{red}{da}-NP. Examples are shown in the Appendix.

\begin{figure}[!h]
\footnotesize
\setlength{\tabcolsep}{3pt} 
\begin{tabular}{lllll} 
\hline
\multicolumn{5}{c}{\sc Caus context}\\
\hline
1 & \textcolor{violet}{Ag} & \textcolor{brown}{Akt} & \textcolor{orange}{Pat} & \textcolor{blue}{P-}NP \\
2 & \textcolor{violet}{Ag} & \textcolor{brown}{Akt} & \textcolor{orange}{Pat} & \textcolor{red}{da}-NP  \\
3 & \textcolor{orange}{Pat} & \textcolor{teal}{Pass} & \textcolor{red}{da}-\textcolor{violet}{Ag} &  \textcolor{blue}{P-}NP \\
4 & \textcolor{orange}{Pat} & \textcolor{teal}{Pass} & \textcolor{red}{da}-\textcolor{violet}{Ag}  & \textcolor{red}{da}-NP\\
5 & \textcolor{orange}{Pat} & \textcolor{teal}{Pass} & & \textcolor{blue}{P}-NP\\
6 & \textcolor{orange}{Pat} & \textcolor{teal}{Pass} &  & \textcolor{red}{da}-NP\\
7 & \textcolor{orange}{Pat} & \textcolor{brown}{Akt} & & \textcolor{blue}{P}-NP\\ 
? & ??? & & &  \\ \hline 
\end{tabular}
\begin{tabular}{ll|l} \hline
\multicolumn{3}{c}{\sc Caus answers}  \\ \hline
1 & \textcolor{orange}{Pat} \textcolor{brown}{Akt}   \textcolor{red}{da}-NP & \textsc{\textbf{Correct}}\\ 
2 & \textcolor{violet}{Ag} \textcolor{brown}{Akt}   \textcolor{red}{da}-NP & \textsc{I-Int}  \\ 
3 & \textcolor{orange}{Pat} \textcolor{teal}{Pass}  \textcolor{red}{da}-\textcolor{violet}{Ag} & \textsc{ER-Pass}\\ 
4 & \textcolor{violet}{Ag} \textcolor{teal}{Pass}   \textcolor{red}{da}-\textcolor{orange}{Pat} & \textsc{IER-Pass}\\ 
5 & \textcolor{orange}{Pat} \textcolor{brown}{Akt}   \textcolor{violet}{Ag} & \textsc{R-Trans}\\ 
6 & \textcolor{violet}{Ag} \textcolor{brown}{Akt}   \textcolor{orange}{Pat} & \textsc{IR-Trans} \\ 
7 & \textcolor{orange}{Pat} \textcolor{brown}{Akt}   \textcolor{red}{da}-\textcolor{violet}{Ag} & \textsc{E-WrBy} \\ 
8 & \textcolor{violet}{Ag} \textcolor{brown}{Akt}   \textcolor{red}{da}-\textcolor{orange}{Pat} & \textsc{IE-WrBy}  \\ 
\hline
\end{tabular}

\begin{tabular}{lllll} 
\hline
\multicolumn{5}{c}{\sc Od context}\\
\hline
1 & \textcolor{violet}{Ag} & \textcolor{brown}{Akt} & \textcolor{orange}{Pat} & \textcolor{blue}{P-}NP\\
2 & \textcolor{violet}{Ag} & \textcolor{brown}{Akt} & \textcolor{orange}{Pat} & \textcolor{red}{da}-NP \\
3 & \textcolor{orange}{Pat} & \textcolor{teal}{Pass} & \textcolor{red}{da}-\textcolor{violet}{Ag}  & \textcolor{blue}{P}-NP\\
4 & \textcolor{orange}{Pat} & \textcolor{teal}{Pass} & \textcolor{red}{da}-\textcolor{violet}{Ag}  & \textcolor{red}{da}-NP\\
5 & \textcolor{orange}{Pat} & \textcolor{teal}{Pass} & & \textcolor{blue}{P}-NP\\
6 & \textcolor{orange}{Pat} & \textcolor{teal}{Pass} &  & \textcolor{red}{da}-NP\\
7 & \textcolor{violet}{Ag} & \textcolor{brown}{Akt} & & \textcolor{blue}{P}-NP\\
? & ??? & & &  \\ \hline 
\end{tabular}
\begin{tabular}{rl|l} \hline
\multicolumn{3}{c}{\sc Od answers}  \\ \hline
1 & \textcolor{orange}{Pat} \textcolor{brown}{Akt}   \textcolor{red}{da}-NP & \textsc{I-Int}\\ 
2 & \textcolor{violet}{Ag} \textcolor{brown}{Akt}   \textcolor{red}{da}-NP & \textsc{\textbf{Correct}}  \\ 
3 & \textcolor{orange}{Pat} \textcolor{teal}{Pass}  \textcolor{red}{da}-\textcolor{violet}{Ag} & \textsc{IER-Pass}\\ 
4 & \textcolor{violet}{Ag} \textcolor{teal}{Pass}   \textcolor{red}{da}-\textcolor{orange}{Pat} & \textsc{ER-Pass}\\ 
5 & \textcolor{orange}{Pat} \textcolor{brown}{Akt}   \textcolor{violet}{Ag} & \textsc{IR-Trans}\\ 
6 & \textcolor{violet}{Ag} \textcolor{brown}{Akt}   \textcolor{orange}{Pat} & \textsc{R-Trans} \\ 
7 & \textcolor{orange}{Pat} \textcolor{brown}{Akt}   \textcolor{red}{da}-\textcolor{violet}{Ag} & \textsc{IE-WrBy} \\ 
8 & \textcolor{violet}{Ag} \textcolor{brown}{Akt}   \textcolor{red}{da}-\textcolor{orange}{Pat} & \textsc{E-WrBy}  \\ 

\hline
\end{tabular}
\caption{BLM contexts answers and their location of errors (see text) for the Change of state group (\caus) and the object drop (\od) class.}
\vspace{-5mm}
\label{tab:templateBLM}
\end{figure}

We illustrate the data in Figure \ref{tab:TypeI-ItCOSfun} in the appendix  with the Italian Change-of-state verb \textit{chiudere} 'close'. 

\paragraph{Lexicalisation} In line with previous work on BLMs, each dataset also contains a varying amount of lexicalisation.  In type I the lexical material of the sentences within a single context does not change, in type II only the verb remains the same, in type III data all words can change (Figure \ref{tab:all-types}, in the appendix).

\subsection{Dataset statistics}

Each subset is split 90:20:10 into train:dev:test subsets. The training and testing are disjoint (agreement data is split based on the correct answer, the alternations data based on the verb). Agreement has  230 test instances for type I, 4121 for types II and III. The verb alternations have 240 test instances for all subsets. We randomly sample a number of training instances, depending on the experimental set-up.

\section{Multi-task representations}
\label{sec:parts-for-BLMs}

\begin{figure}[h]
    \centering
    \includegraphics[width=0.5\textwidth]{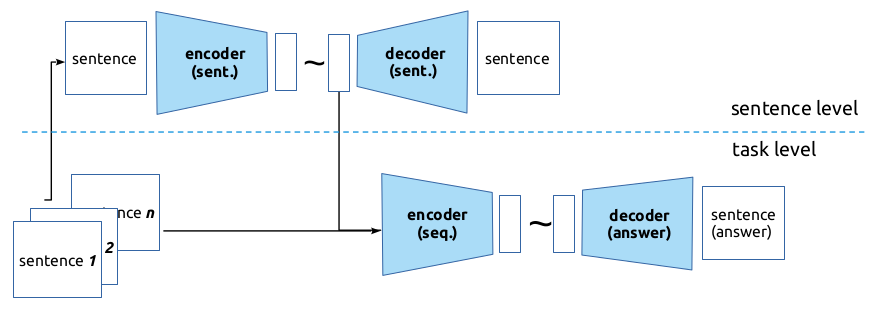}
    \caption{A two-level VAE: the sentence level learns to compress a sentence into a representation useful to solve the BLM problem on the task level.}
    \vspace{-3mm}
    \label{fig:2levelVAE}
\end{figure}

Sentence embeddings encode much information from the input sentence -- lexical, syntactic, semantic, and possibly other types of information. Previous experiments have shown that sentence embeddings can be compressed into very small representations (vectors of size 5) that encode 
information about the structure of the sentence in terms of chunks and their properties, such that they contribute to finding the sequence patterns in BLMs \cite{nastase2024identifiable}. In this work, we investigate whether several BLM tasks 
can share the same structural information from a sentence embedding. Towards this end, we built a multi-task version of a two-level system, illustrated in Figure \ref{fig:2levelVAE}. In this system, one level processes individual sentences and learns to compress them into small vectors that retain information pertinent to a task and the other level uses the compressed sentence representation to find patterns across an input sequence to solve a BLM task. The multi-task variation consists in a single shared sentence-level component, and multiple task components, one for each of the BLM tasks. 
%
%

The BLM problems encode a linguistic phenomenon through data that has structure on multiple levels -- within sentences, and across a sequence of sentences. We can exploit this structure to develop an indirectly supervised approach to discover and use these different levels of structure. We thus model the solving of a BLM task as a two-step process: (i) compress individual sentences into a representation that emphasizes the sentence structure relevant to the BLM problem (e.g. chunks and their grammatical number for the subject-verb agreement task) (ii) use the compressed representations to detect the sequence-level pattern and solve the BLM task. This two-step process has been shown to be used by people solving visual intelligence tests \cite{carpenter1990one}. In our case, this setup allows us to investigate whether the sentence level can be guided to learn shared information, relevant to the different linguistic tasks described in section \ref{sec:data}.

We implement this approach in the two-level intertwined architecture illustrated in Figure \ref{fig:2levelVAE}, and described in detail elsewhere \cite{nastase2024identifiable}. The data is pre-encoded with Electra \cite{clark2020electra}.\footnote{Italian Electra (E-It) pretrained model: dbmdz/electra-base-italian-xxl-cased-discriminator. Multi-lingual Electra (E-M) model: google/electra-base-discriminator.} The sentence representations is provided by the embedding of the [CLS] token.\footnote{To simplify the discussion of the method, we write "sentence" instead of "sentence embedding", when discussing the system.}. We chose Electra because of its stronger sentence-level supervision signal, which leads to higher results when testing the encoding of structural information compared to BERT, RoBERTa, and models tuned by semantic similarity \cite{nastase2024identifiable}.

The two levels are learned together. The input is a BLM instance which is processed on the fly to produce training instances for the sentence level for each sentence $in_k$ in the input sequence $S$. The compressed sentence representations on the latent layer $z_{in_k}$ are stacked and passed as input to the task level, which produces a sentence representation $answ$ as output, which is compared to the answer set of the respective BLM instance $A$.

The sentence level uses a variational encode-decoder architecture to learn how to compress on the latent layer a representation that captures relevant structural information. We guide the system towards this representation by constructing a contrastive set of candidates for comparison with the reconstructed input. The correct output ($out^+$) is the same as the input ($in$), and a selection of other sentences from the input sequence will be the contrastive negative outputs ($Out^- = \{out^-_i, i=1,N_{negs}\}, N_{negs}=7$ (note that an input sequence consists of sentences with different patterns to each other -- Figure \ref{fig:agr_templ} and \ref{tab:templateBLM}). We use a max-margin loss function to take advantage of the contrastive answers, $\hat{in}$ is the reconstructed input sentence from the sampled latent vector $z_{in}$:

\vspace{2mm}
{
\noindent $loss_{sent}(in) = maxM(\hat{in},out^+, Out^-)\\
\mbox{~~~~~~~~~~~~~~~~~~~~~~~~} + KL(z_{in}||\mathcal{N}(0,1))$ \\ 

\noindent $maxM(\hat{i}n,{out^+}, Out^-) = \\ 
\mbox{~~~~~~~~~~~~~~~~~~~~~~~~~~~} max(0,1-cos(\hat{in}, {out^+}) \\
\mbox{~~~~~~~~~~~~~~~~~~~~~~~~~~~}+\frac{\sum_{out^-_i \in Out^-}cos(\hat{in}, {out^-_i})}{N_{negs}})$
}
\vspace{2mm}

The loss at the task level for input sequence $S$ is computed in a similar manner for the constructed answer $answ$, but relative to the answer set $\mathcal{A}$ and the correct answer $a_c$ of the task: \\

\noindent $loss_{task}(S) = maxM(answ, a_c, A\setminus{\{a_c\}}) \\
\mbox{~~~~~~~~~~~~~~~~~~~~~}+ KL_{seq}(z_S|\mathcal{N}(0,1))$. \\

The loss of the two-level systems is: \\

\noindent $loss(S) = \sum_{in_k \in S} loss_{sent}(in_k) + loss_{task}(S)$ \\

The input batches are shuffled, to alternate between tasks during training, and avoid getting stuck in a local maximum for one of the tasks. 

\section{Multi-task results}


Previous published work from our group and current ongoing work has benchmarked the problems generated by some of these datasets \cite{an-etal-2023-blm,nastase-merlo-2023-grammatical}. This work has shown that information about the syntactic phrases in a sentence and their properties can be obtained from sentence embeddings, and this information is helpful in solving the BLM tasks. We had studied these tasks separately, and investigate here whether such structure is encoded in the sentence embeddings, or whether it is assembled based on shallower patterns within the sentence representations.

\begin{figure}[htp]
    \centering
        \includegraphics[width=\linewidth]{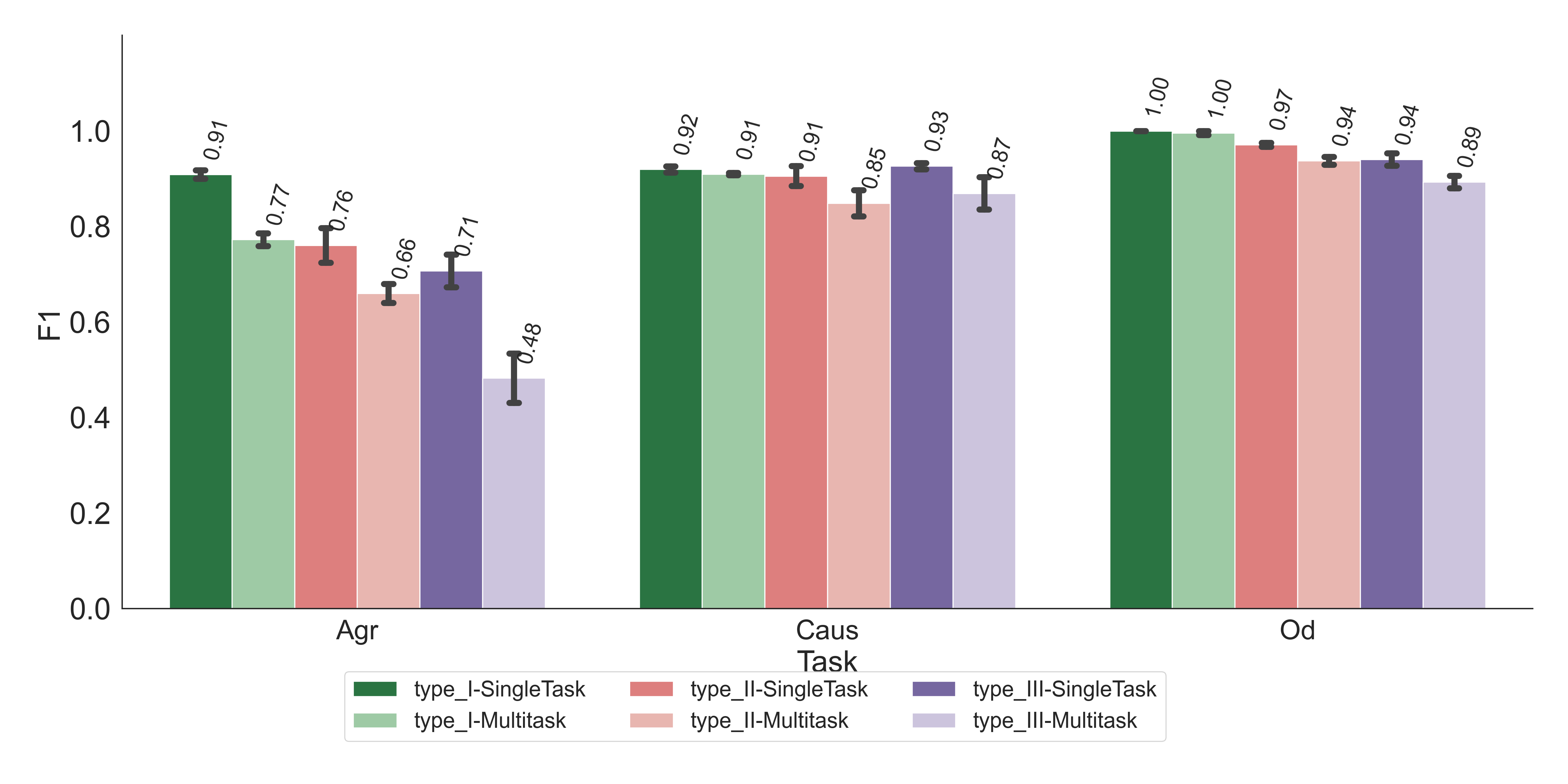}
        \caption{Performance comparison across single-task and multi-task training paradigms for the three subtasks (single task darker shade of each colour, multi-task lighter shade), trained on type-I data, tested on the three types, and averaged over three independent runs. Results obtained using the Italian Electra pretrained model.}
        \vspace{-5mm}
        \label{fig:multitask_vs_single}
\end{figure}

\begin{figure}
    \centering
        \includegraphics[width=0.47\textwidth]{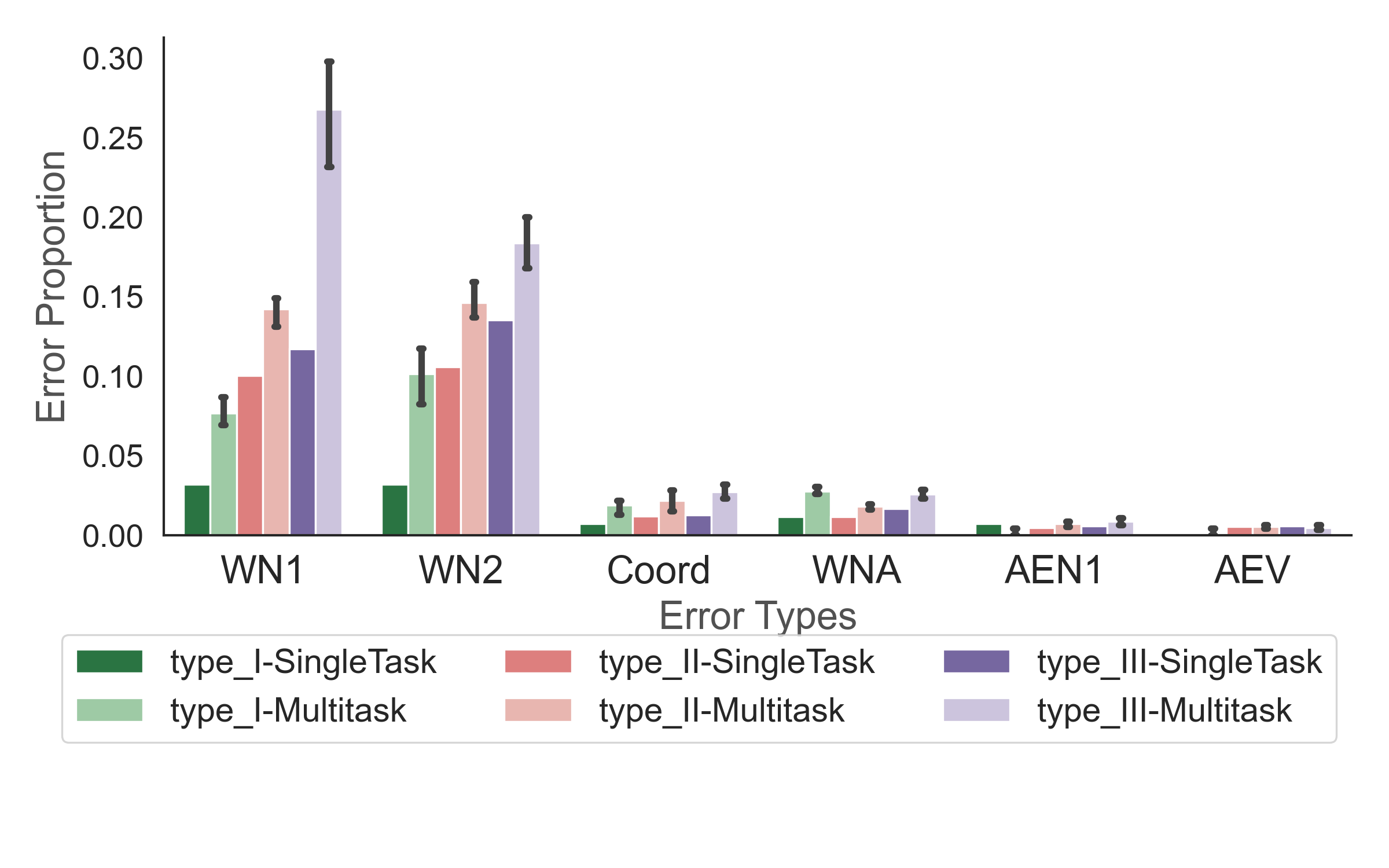}
        \vspace{-7mm}
        \caption{Error analysis for agreement: multi- vs. single task, training on type I data, testing on all.}
        \vspace{-7mm}
        \label{fig:agreement}
\end{figure}

\begin{figure*}[t]
    \begin{minipage}{\textwidth}
    \begin{subfigure}[b]{0.42\textwidth}
        \includegraphics[width=\textwidth,trim={0cm 0cm 0cm 1.5cm},clip]{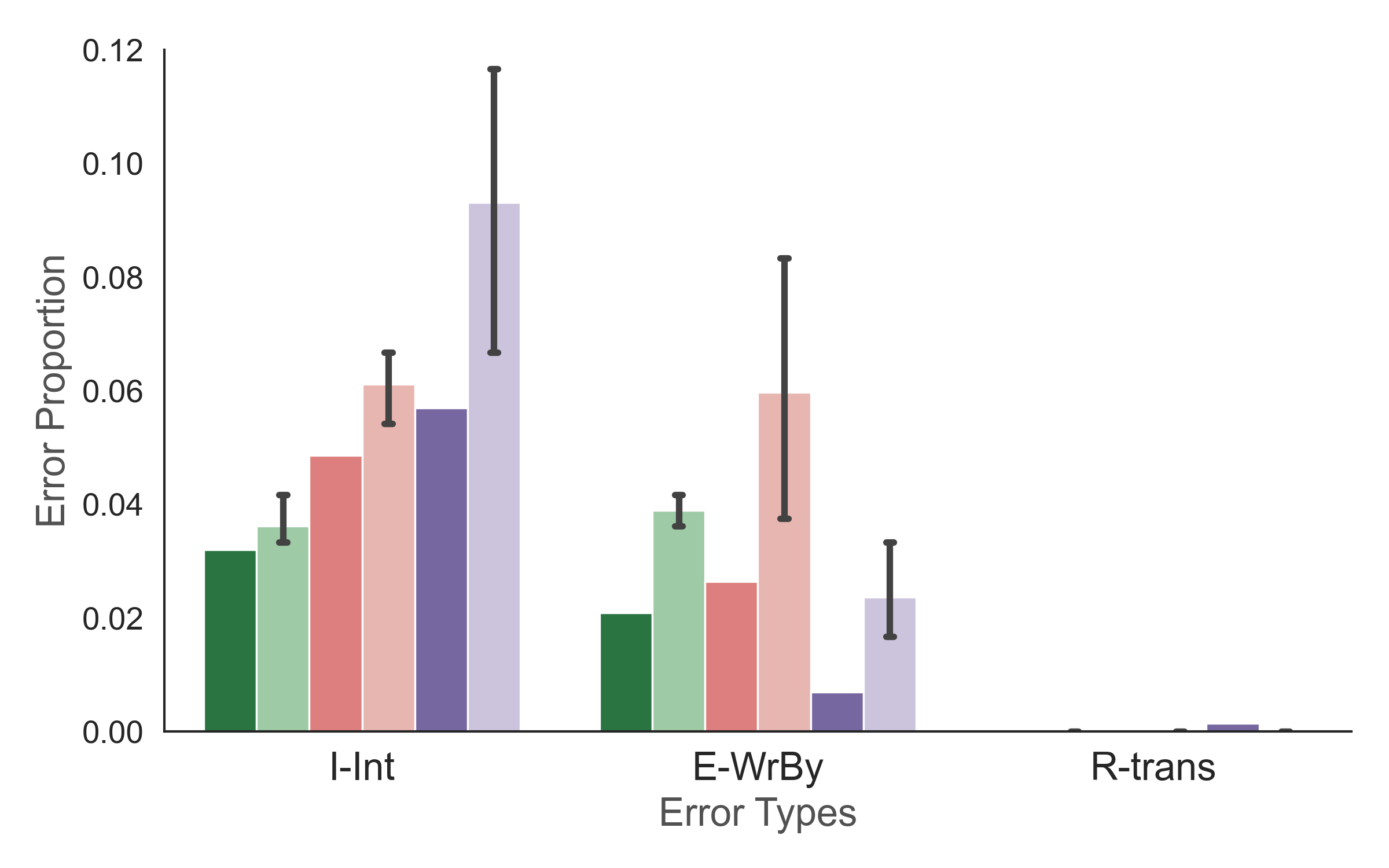}
        \vspace{-7mm}
        \caption{\caus~  task error analysis}
        \label{fig:cos_lex}
    \end{subfigure}
    \hfill
    \begin{subfigure}[b]{0.42\textwidth}
        \includegraphics[width=\textwidth,trim={0cm 0cm 0cm 1.5cm},clip]{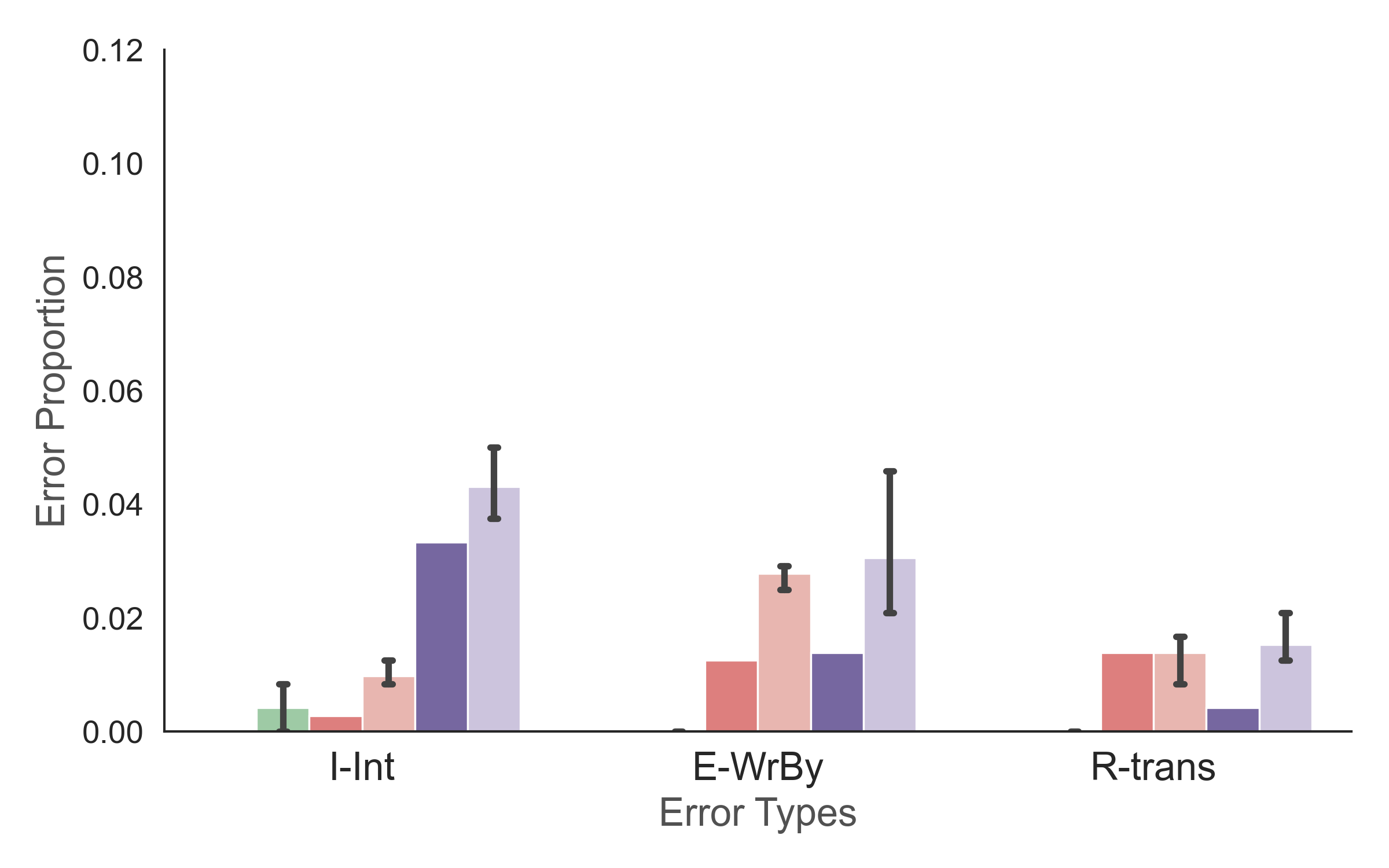}
        \vspace{-7mm}
        \caption{\od~  task error analysis}
        \label{fig:od_lex}
    \end{subfigure}
    \end{minipage}
    \caption{Error analysis between single and multi-task training paradigms trained on type-I data, tested on the three types, as averages over three runs (single task darker shade of each colour, multi-task lighter shade). For the \caus~ and \od~ tasks, we report only three representative error types of \textit{I}, \textit{E} and \textit{R}.}
    \label{fig:errors}
    \vspace{-7mm}
\end{figure*}

\paragraph{Discussion}

We expect that if the multi-task setup succeeds in sharing information across tasks, then the results on the individual test data will be at least as good as when learning tasks individually, given that the multi-task setup uses a larger training set data -- the union of the training sets of the individual tasks. But, overall, this does not seem to be the case. 

As the results in Figure \ref{fig:multitask_vs_single} show  (and also the detailed results in Tables \ref{tab:res_multitask}-\ref{tab:res_singletask} for the Italian Electra pretrained model, and in Tables \ref{tab:res_multitask_EM}-\ref{tab:res_singletask_EM} for a multilingual Electra pretrained model), single-task training outperforms multi-tasking in the agreement and verb alternation subtasks. The drop suggests that the multi-task model is not able to learn shared properties for these tasks, and forcing it to do so leads to a model that is not optimal for either of them. Both tasks require information about the syntactic structure (or sequence of phrases), while each requires different phrase properties -- grammatical number for the agreement task, and semantic properties for the verb alternation. While the system is able to distil all this information from sentence embeddings in the single-task setting, it is not able to compress it into a shared representation when learning the tasks together.

The \od~ single-task and multi-task have comparable performance, probably because the \od~ tasks involve a simpler alternation than the \caus~ task. They do not have a causative meaning and do not require a change in the semantic role of the subjects. 

The comparison of all the tasks suggests that some syntactic and semantic regularities --such as constituents, grammatical number and semantic roles-- cannot be encoded together as they compete with each other when the system learns to distil them from the pretrained sentence embeddings.

\paragraph{Error Analysis}

For the agreement task, errors on the grammatical number of the attractor nouns (\texttt{WN1}, \texttt{WN2}) are high under both paradigms. These are "sequence errors", indicating that the system was not able to detect the patterns in the input sequence, possibly because individual sentence structures were not properly detected. Previous experiments have shown, though, that in the single-task setting, the sentence level does manage to compress the desired information \cite{nastase2024identifiable}. The fact that both these errors increase in the multi-task setting indicates that the information compression on the sentence level is less successful than in the single-task setting.

For the alternation tasks, error patterns vary, although their distributions remain similar between single-task and multi-task environments.

We observe an overall increase of error proportions in the multi-task environment. Specifically, mistakes of the type \textsc{I-int} are frequent in type III data for the \caus~ task. These errors incorrectly map the thematic roles onto the syntax of the arguments (e.g. \textit{L'artista si è chiuso} `the artist closed' or \textit{La carbonara mangiava} `the carbonara was eating'). In the same dataset, we also note an increase of errors related to the last constituent  in type I and type II data (errors of type \textsc{E-WrBy}, e.g. \textit{La finestra si chiuse dall'artista} `the window closed by the artist'). Finally, for the \od~ task, we remark that \textit{R-trans} errors are not the most prominent ---these are the errors resulting in standard transitive clauses (e.g., \textit{L'artista dipinse un paesaggio} `the artist painted a landscape')---  and do not increase in multi-task environments, suggesting that the chosen answer is not derived from some forms of transitive bias \citep{kann-etal-2019-verb}.

An overall comparison shows that the error patterns vary across subtasks. This variety in error patterns confirms that the different dimensions (types of alternations, levels of lexicalisation and single and multi-task learning) are separate uncorrelated dimensions. It also indicates that the differences in the F1 results shown in Figure \ref{fig:multitask_vs_single} are real, despite the more homogeneous trends exhibited by these aggregated F1 numbers.

\section{Conclusions}

In this paper, we have presented curated synthetic datasets of Italian on two linguistic phenomena of an heterogeneous nature, such as agreement and verbal transitive/intransitive alternation, embedded in the BLM task.  

The results on the performance and the error analysis of a tailored two-level architecture have shown that multi-task environments do not help, suggesting that abstract linguistic notions, such as constituents or thematic roles do not seem to be present in the learning process.

Current work is developing new analyses and architectures to probe further in the encoding of information in sentence embeddings and creating new BLM problems across various languages and linguistic phenomena. 

%


\section*{Acknowledgments}

We gratefully acknowledge the partial support of this work by the Swiss National Science Foundation, through grant SNF Advanced grant  TMAG-1\_209426 to PM. 

\bibliographystyle{acl}
\bibliography{bibliography,custom}

\begin{thebibliography}{22}
\expandafter\ifx\csname natexlab\endcsname\relax\def\natexlab#1{#1}\fi
\providecommand{\url}[1]{\texttt{#1}}
\providecommand{\href}[2]{#2}
\providecommand{\path}[1]{#1}
\providecommand{\DOIprefix}{doi:}
\providecommand{\ArXivprefix}{arXiv:}
\providecommand{\URLprefix}{URL: }
\providecommand{\Pubmedprefix}{pmid:}
\providecommand{\doi}[1]{\href{http://dx.doi.org/#1}{\path{#1}}}
\providecommand{\Pubmed}[1]{\href{pmid:#1}{\path{#1}}}
\providecommand{\bibinfo}[2]{#2}
\ifx\xfnm\relax \def\xfnm[#1]{\unskip,\space#1}\fi
\bibitem[{Ruder(2021)}]{ruder2021benchmarking}
\bibinfo{author}{S.~Ruder}, \bibinfo{title}{{Challenges and Opportunities in
  NLP Benchmarking}},
  \bibinfo{howpublished}{\url{http://www.ruder.io/nlp-benchmarking}},
  \bibinfo{year}{2021}.
\bibitem[{Merlo(2023)}]{merlo2023}
\bibinfo{author}{P.~Merlo},
\newblock \bibinfo{title}{Blackbird language matrices {(BLM)}, a new task for
  rule-like generalization in neural networks: Motivations and formal
  specifications},
\newblock \bibinfo{journal}{ArXiv} \bibinfo{volume}{cs.CL 2306.11444}
  (\bibinfo{year}{2023}). \URLprefix
  \url{https://doi.org/10.48550/arXiv.2306.11444}.
  \DOIprefix\doi{10.48550/arXiv.2306.11444}.
\bibitem[{Raven(1938)}]{raven1938}
\bibinfo{author}{J.~C. Raven},
\newblock \bibinfo{title}{Standardization of progressive matrices},
\newblock \bibinfo{journal}{British Journal of Medical Psychology}
  \bibinfo{volume}{19} (\bibinfo{year}{1938}) \bibinfo{pages}{137--150}.
\bibitem[{An et~al.(2023)An, Jiang, A.~Rodriguez, Nastase, and
  Merlo}]{an-etal-2023-blm}
\bibinfo{author}{A.~An}, \bibinfo{author}{C.~Jiang},
  \bibinfo{author}{M.~A.~Rodriguez}, \bibinfo{author}{V.~Nastase},
  \bibinfo{author}{P.~Merlo},
\newblock \bibinfo{title}{{BLM}-{A}gr{F}: A new {F}rench benchmark to
  investigate generalization of agreement in neural networks},
\newblock in: \bibinfo{booktitle}{Proceedings of the 17th Conference of the
  European Chapter of the Association for Computational Linguistics},
  \bibinfo{publisher}{Association for Computational Linguistics},
  \bibinfo{address}{Dubrovnik, Croatia}, \bibinfo{year}{2023}, pp.
  \bibinfo{pages}{1363--1374}. \URLprefix
  \url{https://aclanthology.org/2023.eacl-main.99}.
\bibitem[{Nastase and Merlo(2023)}]{nastase-merlo-2023-grammatical}
\bibinfo{author}{V.~Nastase}, \bibinfo{author}{P.~Merlo},
\newblock \bibinfo{title}{Grammatical information in {BERT} sentence embeddings
  as two-dimensional arrays},
\newblock in: \bibinfo{editor}{B.~Can}, \bibinfo{editor}{M.~Mozes},
  \bibinfo{editor}{S.~Cahyawijaya}, \bibinfo{editor}{N.~Saphra},
  \bibinfo{editor}{N.~Kassner}, \bibinfo{editor}{S.~Ravfogel},
  \bibinfo{editor}{A.~Ravichander}, \bibinfo{editor}{C.~Zhao},
  \bibinfo{editor}{I.~Augenstein}, \bibinfo{editor}{A.~Rogers},
  \bibinfo{editor}{K.~Cho}, \bibinfo{editor}{E.~Grefenstette},
  \bibinfo{editor}{L.~Voita} (Eds.), \bibinfo{booktitle}{Proceedings of the 8th
  Workshop on Representation Learning for NLP (RepL4NLP 2023)},
  \bibinfo{publisher}{Association for Computational Linguistics},
  \bibinfo{address}{Toronto, Canada}, \bibinfo{year}{2023}, pp.
  \bibinfo{pages}{22--39}. \URLprefix
  \url{https://aclanthology.org/2023.repl4nlp-1.3}.
  \DOIprefix\doi{10.18653/v1/2023.repl4nlp-1.3}.
\bibitem[{Nastase and Merlo(2024)}]{nastase2024identifiable}
\bibinfo{author}{V.~Nastase}, \bibinfo{author}{P.~Merlo},
\newblock \bibinfo{title}{Are there identifiable structural parts in the
  sentence embedding whole?},
\newblock in: \bibinfo{editor}{Y.~Belinkov}, \bibinfo{editor}{N.~Kim},
  \bibinfo{editor}{J.~Jumelet}, \bibinfo{editor}{H.~Mohebbi},
  \bibinfo{editor}{A.~Mueller}, \bibinfo{editor}{H.~Chen} (Eds.),
  \bibinfo{booktitle}{Proceedings of the 7th BlackboxNLP Workshop: Analyzing
  and Interpreting Neural Networks for NLP}, \bibinfo{publisher}{Association
  for Computational Linguistics}, \bibinfo{address}{Miami, Florida, US},
  \bibinfo{year}{2024}, pp. \bibinfo{pages}{23--42}. \URLprefix
  \url{https://aclanthology.org/2024.blackboxnlp-1.3}.
\bibitem[{Lenci(2023)}]{lenci2023NLU}
\bibinfo{author}{A.~Lenci},
\newblock \bibinfo{title}{Understanding natural language understanding
  systems},
\newblock \bibinfo{journal}{Sistemi intelligenti, Rivista quadrimestrale di
  scienze cognitive e di intelligenza artificiale}  (\bibinfo{year}{2023})
  \bibinfo{pages}{277--302}. \URLprefix
  \url{https://www.rivisteweb.it/doi/10.1422/107438}.
  \DOIprefix\doi{10.1422/107438}.
\bibitem[{Carpenter et~al.(1990)Carpenter, Just, and Shell}]{carpenter-ea1990}
\bibinfo{author}{P.~A. Carpenter}, \bibinfo{author}{M.~A. Just},
  \bibinfo{author}{P.~Shell},
\newblock \bibinfo{title}{What one intelligence test measures: A theoretical
  account of the processing in the {R}aven {P}rogressive {M}atrices {T}est.},
\newblock \bibinfo{journal}{Psychological Review} \bibinfo{volume}{97}
  (\bibinfo{year}{1990}) \bibinfo{pages}{404--431}.
  \DOIprefix\doi{10.1037/0033-295X.97.3.404}.
\bibitem[{Zhang et~al.(2023)Zhang, Yu, Yu, Guo, and
  Jiang}]{zhang-etal-2023-survey}
\bibinfo{author}{Z.~Zhang}, \bibinfo{author}{W.~Yu}, \bibinfo{author}{M.~Yu},
  \bibinfo{author}{Z.~Guo}, \bibinfo{author}{M.~Jiang},
\newblock \bibinfo{title}{A survey of multi-task learning in natural language
  processing: Regarding task relatedness and training methods},
\newblock in: \bibinfo{editor}{A.~Vlachos}, \bibinfo{editor}{I.~Augenstein}
  (Eds.), \bibinfo{booktitle}{Proceedings of the 17th Conference of the
  European Chapter of the Association for Computational Linguistics},
  \bibinfo{publisher}{Association for Computational Linguistics},
  \bibinfo{address}{Dubrovnik, Croatia}, \bibinfo{year}{2023}, pp.
  \bibinfo{pages}{943--956}. \URLprefix
  \url{https://aclanthology.org/2023.eacl-main.66}.
  \DOIprefix\doi{10.18653/v1/2023.eacl-main.66}.
\bibitem[{Chen et~al.(2021)Chen, Zhang, and Yang}]{chen2021multi}
\bibinfo{author}{S.~Chen}, \bibinfo{author}{Y.~Zhang},
  \bibinfo{author}{Q.~Yang},
\newblock \bibinfo{title}{Multi-task learning in natural language processing:
  An overview},
\newblock \bibinfo{journal}{ACM Computing Surveys}  (\bibinfo{year}{2021}).
\bibitem[{Ruder(2017)}]{ruder2017overview}
\bibinfo{author}{S.~Ruder},
\newblock \bibinfo{title}{An overview of multi-task learning in deep neural
  networks},
\newblock \bibinfo{journal}{arXiv preprint arXiv:1706.05098}
  (\bibinfo{year}{2017}).
\bibitem[{Pfeiffer et~al.(2023)Pfeiffer, Ruder, Vuli{\'c}, and
  Ponti}]{pfeiffer2023modular}
\bibinfo{author}{J.~Pfeiffer}, \bibinfo{author}{S.~Ruder},
  \bibinfo{author}{I.~Vuli{\'c}}, \bibinfo{author}{E.~M. Ponti},
\newblock \bibinfo{title}{Modular deep learning},
\newblock \bibinfo{journal}{arXiv preprint arXiv:2302.11529}
  (\bibinfo{year}{2023}).
\bibitem[{Standley et~al.(2020)Standley, Zamir, Chen, Guibas, Malik, and
  Savarese}]{standley2020tasks}
\bibinfo{author}{T.~Standley}, \bibinfo{author}{A.~Zamir},
  \bibinfo{author}{D.~Chen}, \bibinfo{author}{L.~Guibas},
  \bibinfo{author}{J.~Malik}, \bibinfo{author}{S.~Savarese},
\newblock \bibinfo{title}{Which tasks should be learned together in multi-task
  learning?},
\newblock in: \bibinfo{booktitle}{International conference on machine
  learning}, \bibinfo{organization}{PMLR}, \bibinfo{year}{2020}, pp.
  \bibinfo{pages}{9120--9132}.
\bibitem[{Zhou et~al.(2021)Zhou, Cai, Zhang, and Yuan}]{zhou-etal-2021-end}
\bibinfo{author}{B.~Zhou}, \bibinfo{author}{X.~Cai},
  \bibinfo{author}{Y.~Zhang}, \bibinfo{author}{X.~Yuan},
\newblock \bibinfo{title}{An end-to-end progressive multi-task learning
  framework for medical named entity recognition and normalization},
\newblock in: \bibinfo{editor}{C.~Zong}, \bibinfo{editor}{F.~Xia},
  \bibinfo{editor}{W.~Li}, \bibinfo{editor}{R.~Navigli} (Eds.),
  \bibinfo{booktitle}{Proceedings of the 59th Annual Meeting of the Association
  for Computational Linguistics and the 11th International Joint Conference on
  Natural Language Processing (Volume 1: Long Papers)},
  \bibinfo{publisher}{Association for Computational Linguistics},
  \bibinfo{address}{Online}, \bibinfo{year}{2021}, pp.
  \bibinfo{pages}{6214--6224}. \URLprefix
  \url{https://aclanthology.org/2021.acl-long.485}.
  \DOIprefix\doi{10.18653/v1/2021.acl-long.485}.
\bibitem[{Hu et~al.(2022)Hu, Chan, and Huang}]{hu-etal-2022-mocha}
\bibinfo{author}{Z.~Hu}, \bibinfo{author}{H.~P. Chan},
  \bibinfo{author}{L.~Huang},
\newblock \bibinfo{title}{{MOCHA}: A multi-task training approach for coherent
  text generation from cognitive perspective},
\newblock in: \bibinfo{editor}{Y.~Goldberg}, \bibinfo{editor}{Z.~Kozareva},
  \bibinfo{editor}{Y.~Zhang} (Eds.), \bibinfo{booktitle}{Proceedings of the
  2022 Conference on Empirical Methods in Natural Language Processing},
  \bibinfo{publisher}{Association for Computational Linguistics},
  \bibinfo{address}{Abu Dhabi, United Arab Emirates}, \bibinfo{year}{2022}, pp.
  \bibinfo{pages}{10324--10334}. \URLprefix
  \url{https://aclanthology.org/2022.emnlp-main.705}.
  \DOIprefix\doi{10.18653/v1/2022.emnlp-main.705}.
\bibitem[{Wang et~al.(2020)Wang, Zhai, and Hassan}]{wang-etal-2020-multi}
\bibinfo{author}{Y.~Wang}, \bibinfo{author}{C.~Zhai},
  \bibinfo{author}{H.~Hassan},
\newblock \bibinfo{title}{Multi-task learning for multilingual neural machine
  translation},
\newblock in: \bibinfo{editor}{B.~Webber}, \bibinfo{editor}{T.~Cohn},
  \bibinfo{editor}{Y.~He}, \bibinfo{editor}{Y.~Liu} (Eds.),
  \bibinfo{booktitle}{Proceedings of the 2020 Conference on Empirical Methods
  in Natural Language Processing (EMNLP)}, \bibinfo{publisher}{Association for
  Computational Linguistics}, \bibinfo{address}{Online}, \bibinfo{year}{2020},
  pp. \bibinfo{pages}{1022--1034}. \URLprefix
  \url{https://aclanthology.org/2020.emnlp-main.75}.
  \DOIprefix\doi{10.18653/v1/2020.emnlp-main.75}.
\bibitem[{Devlin et~al.(2019)Devlin, Chang, Lee, and
  Toutanova}]{devlin-etal-2019-bert}
\bibinfo{author}{J.~Devlin}, \bibinfo{author}{M.-W. Chang},
  \bibinfo{author}{K.~Lee}, \bibinfo{author}{K.~Toutanova},
\newblock \bibinfo{title}{{BERT}: Pre-training of deep bidirectional
  transformers for language understanding},
\newblock in: \bibinfo{booktitle}{Proceedings of the 2019 Conference of the
  North {A}merican Chapter of the Association for Computational Linguistics:
  Human Language Technologies, Volume 1 (Long and Short Papers)},
  \bibinfo{publisher}{Association for Computational Linguistics},
  \bibinfo{address}{Minneapolis, Minnesota}, \bibinfo{year}{2019}, pp.
  \bibinfo{pages}{4171--4186}. \URLprefix
  \url{https://aclanthology.org/N19-1423}.
  \DOIprefix\doi{10.18653/v1/N19-1423}.
\bibitem[{Clark et~al.(2020)Clark, Luong, Le, and Manning}]{clark2020electra}
\bibinfo{author}{K.~Clark}, \bibinfo{author}{M.-T. Luong},
  \bibinfo{author}{Q.~V. Le}, \bibinfo{author}{C.~D. Manning},
\newblock \bibinfo{title}{Electra: Pre- training text encoders as
  discriminators rather than generators},
\newblock in: \bibinfo{booktitle}{ICLR}, \bibinfo{year}{2020}, pp.
  \bibinfo{pages}{1--18}.
\bibitem[{Levin(1993)}]{Levin93}
\bibinfo{author}{B.~Levin}, \bibinfo{title}{{English Verb Classes and
  Alternations A Preliminary Investigation}}, \bibinfo{publisher}{University of
  Chicago Press}, \bibinfo{address}{Chicago and London}, \bibinfo{year}{1993}.
\bibitem[{Merlo and Stevenson(2001)}]{merlo2001automatic}
\bibinfo{author}{P.~Merlo}, \bibinfo{author}{S.~Stevenson},
\newblock \bibinfo{title}{Automatic verb classification based on statistical
  distributions of argument structure},
\newblock \bibinfo{journal}{Computational Linguistics} \bibinfo{volume}{27}
  (\bibinfo{year}{2001}) \bibinfo{pages}{373--408}.
\bibitem[{Carpenter et~al.(1990)Carpenter, Just, and Shell}]{carpenter1990one}
\bibinfo{author}{P.~A. Carpenter}, \bibinfo{author}{M.~A. Just},
  \bibinfo{author}{P.~Shell},
\newblock \bibinfo{title}{What one intelligence test measures: a theoretical
  account of the processing in the raven progressive matrices test.},
\newblock \bibinfo{journal}{Psychological review} \bibinfo{volume}{97}
  (\bibinfo{year}{1990}) \bibinfo{pages}{404}.
\bibitem[{Kann et~al.(2019)Kann, Warstadt, Williams, and
  Bowman}]{kann-etal-2019-verb}
\bibinfo{author}{K.~Kann}, \bibinfo{author}{A.~Warstadt},
  \bibinfo{author}{A.~Williams}, \bibinfo{author}{S.~R. Bowman},
\newblock \bibinfo{title}{Verb argument structure alternations in word and
  sentence embeddings},
\newblock in: \bibinfo{booktitle}{Proceedings of the Society for Computation in
  Linguistics ({SC}i{L}) 2019}, \bibinfo{year}{2019}, pp.
  \bibinfo{pages}{287--297}. \URLprefix
  \url{https://aclanthology.org/W19-0129}. \DOIprefix\doi{10.7275/q5js-4y86}.

\end{thebibliography}

\newpage
\onecolumn
\appendix
\section{Appendix}

\subsection{An Italian example for the subject-verb agreement BLM}
\begin{figure}[h]
\centering
\small
\setlength{\tabcolsep}{5pt}
\begin{tabular}{lllll} \hline
 \multicolumn{5}{c}{\sc Context}  \\ \hline
1 &  \textcolor{blue}{Il vaso}   & \textcolor{blue}{con il fiore}  &                                     & \textcolor{blue}{si \`{e} rotto.}  \\
2 &  \textcolor{red}{I vasi} & \textcolor{blue}{con il fiore}  &                                     & \textcolor{red}{si sono rotti.}  \\
3 &  \textcolor{blue}{Il vaso}   & \textcolor{red}{con i fiori} &                                     & \textcolor{blue}{si \`{e} rotto.}  \\
4 &  \textcolor{red}{I vasi} & \textcolor{red}{con i fiori} &                                     & \textcolor{red}{si sono rotti.}  \\
5 & \textcolor{blue}{Il vaso}   & \textcolor{blue}{con il fiore}  & \textcolor{blue}{del giardino}   & \textcolor{blue}{si \`{e} rotto.}  \\
6 & \textcolor{red}{I vasi} & \textcolor{blue}{con il fiore}  & \textcolor{blue}{del giardino}  & \textcolor{red}{si sono rotti.}  \\
7 & \textcolor{blue}{Il vaso}   & \textcolor{red}{con i fiori} & \textcolor{blue}{del giardino}  & \textcolor{blue}{si \`{e} rotto.}  \\
8 &???
\end{tabular}
 \\ 
 \setlength{\tabcolsep}{2pt}
   \begin{tabular}{rll} \hline
 \multicolumn{3}{c}{\sc Answer set}  \\ \hline
 1 &Il vaso con i fiori e il giardino si \`{e} rotto. & coord  \\ 
 2 & {\bf I vasi con i fiori del giardino si sono rotti.} & correct \\
 3 & Il vaso con il fiore si \`{e} rotto. &  WNA\\
 4 & I vasi con il fiore del giardino si sono rotti.
& WN1\\
 5 & I vasi con i fiori dei giardini si sono rotti. & WN2 \\ 
 6 & Il vaso con il fiore del giardino si sono rotti. & AEV \\
 7 & Il vaso con i fiori del giardino si sono rotti. & AEN1 \\
 8 & Il vaso con il fiore dei giardini si sono rotti. & AEN2 \\\hline
 \end{tabular}
\caption{An illustrative example for the BLM instances for verb-subject agreement, with 2 attractors (\textit{fiore} 'flower', \textit{giardino} 'garden'), with  candidate answer set. }
\vspace{-5mm}
\label{fig:matrix-example}
\end{figure}

\subsection{Verb alternation examples}

\begin{figure*}[!h]
\centering
\footnotesize
\setlength{\tabcolsep}{5pt} 

\begin{tabular}{|p{0.1cm}|p{0.42\textwidth}|} 
\hline
\multicolumn{2}{|c|}{\sc Caus - Context} \\
\hline
1 & Una stella del cinema chiuse la sua carriera con forza \\
2 & Una stella del cinema chiuse la sua carriera da pochissimo tempo \\
3 & La sua carriera fu chiusa da una stella del cinema con forza\\
4 & La sua carriera fu chiusa da una stella del cinema da pochissimo tempo\\
5 & La sua carriera fu chiusa con forza \\
6 & La sua carriera fu chiusa da pochissimo tempo \\
7 & La sua carriera si chiuse con forza \\ 
8 & ??? \\ 
\hline 
\end{tabular} 
\begin{tabular}{|p{0.1cm}|p{0.42\textwidth}|} 
\hline
\multicolumn{2}{|c|}{\sc Caus - Answers} \\
\hline
1 & \textbf{La sua carriera si chiuse da pochissimo tempo} \\
2 & Una stella del cinema si chiuse da pochissimo tempo \\
3 & La sua carriera fu chiusa da una stella del cinema \\
4 & Una stella del cinema fu chiusa dalla sua carriera \\
5 & La sua carriera chiuse una stella del cinema \\
6 & Una stella del cinema chiuse la sua carriera \\
7 & La sua carriera si chiuse da una stella del cinema \\
8 & Una stella del cinema si chiuse dalla sua carriera \\
\hline
\end{tabular}
\caption{Examples for the \caus~ BLMs for the Italian verb \textit{chiudere} 'close' belonging to \caus~ class}
\label{tab:TypeI-ItCOSfun}
\end{figure*}

\begin{figure*}[!h]
\centering
\footnotesize
\setlength{\tabcolsep}{5pt} 
\begin{tabular}{cc} %
\begin{tabular}{|p{0.1cm}|p{0.42\textwidth}|} 
\hline
\multicolumn{2}{|c|}{\sc Od, typeI - Context} \\
\hline
1 & La turista mangia una carbonara in un secondo \\
2 & La turista mangia una carbonara da mezz'ora \\
3 & Una carbonara è mangiata dalla turista in un secondo \\
4 & Una carbonara è mangiata dalla turista da mezz'ora\\
5 & Una carbonara è mangiata in un secondo \\
6 & Una carbonara è mangiata da mezz'ora\\
7 & La turista mangia in un secondo\\
8 & ??? \\ \hline 
\end{tabular} 
&
\begin{tabular}{|p{0.1cm}|p{0.42\textwidth}|} 
\hline
\multicolumn{2}{|c|}{\sc Od, typeI  - Answers} \\
\hline
1 & Una carbonara mangia da mezz'ora \\
2 & \textbf{La turista mangia da mezz'ora} \\
3 & Una carbonara è mangiata dalla turista \\
4 & La turista è mangiata da una carbonara\\
5 & Una carbonara mangia la turista\\
6 & La turista mangia una carbonara\\
7 & Una carbonara mangia dalla turista\\
8 & La turista mangia da una carbonara\\
\hline
\end{tabular}
\\ \\ \\\hline\\\\
\begin{tabular}{|p{0.1cm}|p{0.42\textwidth}|} 
\hline
\multicolumn{2}{|c|}{\sc Od, typeII - Context} \\
\hline
1 & La zia mangia una bistecca nella sala grande \\
2 & La presidente può mangiare una bistecca da programma \\
3 & La specialità della casa deve essere mangiata dalla turista nella sala grande\\
4 & Una bistecca fu mangiata dalla presidente da sola\\
5 & La specialità della casa deve essere mangiata in un secondo \\
6 & Una bistecca deve poter essere mangiata da sola\\
7 & La turista deve mangiare con fame\\ 
8 & ??? \\ 
\hline 
\end{tabular} 
&
\begin{tabular}{|p{0.1cm}|p{0.42\textwidth}|} 
\hline
\multicolumn{2}{|c|}{\sc Od, typeII - Answers} \\
\hline
1 & La specialità della casa può mangiare da sola \\
2 & \textbf{La squadra di calcio deve mangiare da mezz'ora}\\
3 & Una bistecca è mangiata dalla turista\\
4 & La squadra di calcio può essere mangiata da una carbonara\\
5 & La pasta col pomodoro può mangiare la squadra di calcio\\
6 & La squadra di calcio mangia una bistecca \\
7 & La specialità della casa deve poter mangiare dalla turista\\
8 & La presidente mangia da una bistecca\\
\hline
\end{tabular}
\\ \\ \\\hline\\\\
\begin{tabular}{|p{0.1cm}|p{0.42\textwidth}|} 
\hline
\multicolumn{2}{|c|}{\sc Od, typeIII - Context} \\
\hline
1 & L'attore deve canticchiare un motivetto dopo il festival \\
2 & L'amica di mia mamma deve cucire la tasca da qualche giorno\\
3 & L'inno nazionale può essere cantato dal vincitore del festival con solo pianoforte\\
4 & Una bistecca deve essere mangiata dalla turista da sola\\
5 & Il manuale è insegnato nell'aula magna\\
6 & Questi attrezzi devono essere intagliati da manuale\\
7 & I due fratelli studiano con molta attenzione\\ 
8 & ??? \\ 
\hline 
\end{tabular} 
&
\begin{tabular}{|p{0.1cm}|p{0.42\textwidth}|} 
\hline
\multicolumn{2}{|c|}{\sc Od, typeIII - Answers} \\
\hline
1 & La pasta frolla deve impastare da sola \\
2 & \textbf{L'autrice deve poter scrivere da qualche giorno}\\
3 & I libri di testo devono poter essere studiati dai candidati \\
4 & Questi stilisti devono poter essere tessuti dai vestiti per la parata\\
5 & Questi motivi greci possono tessere questi stilisti \\
6 & L'idraulico saldò i cavi del lampadario \\
7 & La stanza pulisce da una delle propretarie dell'albergo \\
8 & Le sommozzatrici pescarono da delle trote \\
\hline
\end{tabular}
\end{tabular}
\caption{Examples of \od~ BLMs for type I, type II and type III}
\label{tab:all-types}
\end{figure*}

\clearpage
\section{Results}

\subsection{Results with the Italian Electra pretrained model: \\ dbmdz/electra-base-
italian-xxl-cased-discriminator}

\begin{table}[h!]
  \centering
    \begin{tabular}{ll|ccc}
   {\bf train on} & {\bf test on} & \multicolumn{3}{c}{\bf task} \\ \cline{3-5}		
   &  & agreement	& \caus 	&	\od 	\\ \hline
type I	& type I   & {\bf 0.772 (0.011)}  & 0.910 (0.002) &		{\bf 0.996 (0.003)} \\ 
	    & type II  & {\bf 0.660 (0.016)}  &	0.849 (0.022) &		0.938 (0.007) \\
	    & type III & {\bf 0.483 (0.042)}  &	0.870 (0.027) &		0.893 (0.010) \\ \hline
type II	& type I   & 0.504 (0.056) &	  0.917 (0.012)   & 0.993 (0.004) \\
	      & type II  & 0.519 (0.027) &  	0.872 (0.007)   & 0.981 (0.007) \\
	    & type III & 0.406 (0.018) &	  {\bf 0.907 (0.004)}   & 0.950 (0.009) \\ \hline
type III & type I  & 0.274 (0.012) &		{\bf 0.946 (0.003)} &	0.994 (0.002) \\ 
	    & type II  & 0.330 (0.004) &		{\bf 0.929 (0.003)} &	{\bf 0.983 (0.003)} \\
 	    & type III & 0.325 (0.008) &		0.889 (0.014) &		{\bf 0.967 (0.007)} \\ \hline
    \end{tabular}
    \caption{{\bf Multi-task} learning results as F1 averages over three runs (and standard deviation). Training with 3000 instances -- 1000 from each task.}
    \label{tab:res_multitask}
\end{table}

\begin{table}[h!]
    \begin{tabular}{ll|ccc}
   {\bf train on} & {\bf test on} & \multicolumn{3}{c}{\bf task} \\ \cline{3-5}		
   &  & agreement	& \caus 	&	\od 	\\ \hline
type I	& type I   & {\bf 0.909 (0.007)}  &      0.919 (0.005) &	{\bf 1.000 (0.000)} \\ 
	    & type II  & {\bf 0.760 (0.030)}  &	     0.906 (0.017) &	 	 0.971 (0.003) \\
	    & type III & {\bf 0.707 (0.028)}  &	     0.926 (0.005) &		 0.940 (0.010) \\ \hline
type II	& type I   &      0.881 (0.013)   &	     0.932 (0.007) &	     1.000 (0.000) \\
	      & type II  &      0.784 (0.007)   &      0.903 (0.010) &	       0.983 (0.003) \\
	    & type III &      0.714 (0.005)   &	{\bf 0.956 (0.005)} &	     0.975 (0.009) \\ \hline
type III & type I  &      0.296 (0.011)   &	{\bf 0.960 (0.005)} &	     0.998 (0.002) \\ 
	    & type II  &      0.345 (0.002)   &	{\bf 0.950 (0.007)} &	{\bf 0.993 (0.004)} \\
 	    & type III &      0.336 (0.005)   &	      0.918 (0.010) &	{\bf 0.994 (0.004)} \\ \hline
    \end{tabular}
    \caption{{\bf Single task} learning results as F1 averages over three runs (and standard deviation). Training with 2160 instances for \caus~ and \od~ for all types, and for agreement 2052 instances for type I (maximum available), and 3000 instances for type II and type III.}
    \label{tab:res_singletask}
\end{table}

\newpage 
\subsection{Results with the multilingual Electra pretrained model: \\ google/electra-base-discriminator}

\begin{table}[h!]
  \centering
    \begin{tabular}{ll|ccc}
   {\bf train on} & {\bf test on} & \multicolumn{3}{c}{\bf task} \\ \cline{3-5}		
   &  & agreement	& \caus 	&	\od 	\\ \hline
type I	& type I   &  {\bf 0.664 (0.053)} & 0.543 (0.011)  & 0.714 (0.012) \\ 
	    & type II  &  {\bf 0.733 (0.018)} & 0.407 (0.023)  & 0.561 (0.002) \\
	    & type III &  {\bf 0.586 (0.022)} & 0.483 (0.016)  & 0.656 (0.016) \\ \hline
type II	& type I   &  0.599 (0.025) & {\bf 0.610 (0.035)}  & 0.646 (0.010) \\
	      & type II  &  0.660 (0.019) & {\bf 0.536 (0.004)}  & 0.601 (0.004) \\
	    & type III &  0.518 (0.025) & {\bf 0.601 (0.011)}  & {\bf 0.686 (0.019)} \\ \hline
type III & type I  &  0.320 (0.047) & 0.551 (0.014)  & {\bf 0.729 (0.015)} \\ 
	    & type II  &  0.401 (0.058) & 0.450 (0.021)  & {\bf 0.661 (0.020)}\\
 	    & type III &  0.378 (0.052) & 0.413 (0.012)  & 0.618 (0.005) \\ \hline
    \end{tabular}
    \caption{{\bf Multi-task} learning results as F1 averages over three runs (and standard deviation). Training with 3000 instances -- 1000 from each task.}
    \label{tab:res_multitask_EM}
\end{table}

\begin{table}[h!]
    \begin{tabular}{ll|ccc}
   {\bf train on} & {\bf test on} & \multicolumn{3}{c}{\bf task} \\ \cline{3-5}		
   &  & agreement	& \caus 	&	\od 	\\ \hline
type I	& type I   &  {\bf 0.875 (0.031)}  & 0.599 (0.040) &	0.749 (0.030) \\ 
	    & type II  &  {\bf 0.886 (0.005)} & 0.425 (0.019)  & 0.579 (0.037) \\
	    & type III &  0.815 (0.016) & 0.529 (0.020) & 0.660 (0.014)	 \\ \hline
type II	& type I   &  0.841 (0.024) & 0.543 (0.027)  & 0.651 (0.007) \\
	      & type II  &  0.881 (0.003) & 0.486 (0.005) & 0.596 (0.010)\\
	    & type III &  0.814 (0.008) & {\bf 0.582 (0.026)} & {\bf 0.685 (0.013)} \\ \hline
type III & type I  &  0.826 (0.022) & {\bf 0.632 (0.023)} & {\bf 0.761 (0.023)} \\ 
	    & type II  &  0.878 (0.005) & {\bf 0.557 (0.013)} & {\bf 0.697 (0.009)}	 \\
 	    & type III &  {\bf 0.874 (0.006)} & 0.475 (0.010) & 0.592 (0.024) \\ \hline
    \end{tabular}
    \caption{{\bf Single task} learning results as F1 averages over three runs (and standard deviation). Training with 2160 instances for \caus~ and \od~ for all types, and for agreement 2052 instances for type I (maximum available), and 3000 instances for type II and type III.}
    \label{tab:res_singletask_EM}
\end{table}

\end{document}